# Image Registration Techniques: A Survey

[1]Sayan Nag[*]
[1]Department of Electrical Engineering
Jadavpur University

*Abstract*— Image Registration is the process of aligning two or more images of the same scene with reference to a particular image. The images are captured from various sensors at different times and at multiple view-points. Thus to get a better picture of any change of a scene/object over a considerable period of time image registration is important. Image registration finds application in medical sciences, remote sensing and in computer vision. This paper presents a detailed review of several approaches which are classified accordingly along with their contributions and drawbacks. The main steps of an image registration procedure are also discussed. Different performance measures are presented that determine the registration quality and accuracy. The scope for the future research are presented as well.

## I. Introduction

Image Registration is interpreted as the process of overlaying two or more images of the same scene with respect to a particular reference image. The images may be taken at various circumstances (time-points), from various perspectives (view-points), and additionally by various sensors. The reference image is generally one of these captured images. It geometrically transforms different sets of data into a particular reference co-ordinate system. The discrepancies among these images are interposed owing to the disparate imaging conditions. Image acquisition devices underwent rapid modifications and proliferating amount and diversity of acquired images elicited the research on automatic image registration.

In image analysis ventures, one of the most significant step is Image Registration. It is a necessary step to obtain the final information from a combination of a multitude of divergent sources capturing the same information in varied circumstances and diverse manners. Essentially the objective is to detect the concealed relationship existing between the input and the reference images which is usually indicated by a coordinate transformation matrix. Accordingly, an image registration can be essentially devised as an optimization problem. Image registration plays a crucial role in many real-world applications.

Image registration finds applications in remote sensing [1-3] involving multispectral classification, environmental monitoring, change detection, image mosaicing, weather forecasting, creating super-resolution images and integrating information into geographic information systems (GIS), in medicine [4-8] including fusion of computer tomography (CT) and NMR data to obtain more complete information about the patient, multi-modal analysis of different diseases like epilepsy where the protocols incorporate functional EEG/MEG data along with anatomical MRI, monitoring tumor evolution, treatment verification, juxtaposition of the patient's data with anatomical atlases, in cartography for map updating, and in computer vision for target localization, automatic quality control and motion tracking.

According to the manner of image acquisition the application of Image Registration can be segregated into the following groups.

1. *Multi-view Analysis*: Images of the similar object or scene are captured from multiple viewpoints to gain a better representation of the scanned object or scene. Examples include mosaicing of images and shape recovery from the stereo.

2. *Multi-temporal Analysis*: Images of the same object/scene are captured at various times usually under dissimilar conditions to notice changes in the object/scene which emerged between the successive images acquisitions. Examples include motion tracking, tracking the growth of tumors.

3. *Multi-modal Analysis*: Different sensors are used to acquire the images of the same object/scene to merge the information obtained from various sources to obtain the minutiae of the object/scene. Examples include integration of information from sensors with disparate characteristics providing better spatial and spectral resolutions independent of illumination-this depends upon the robustness of the registration algorithm, combination of sensors capturing the anatomical information like magnetic resonance image (MRI), ultrasound or CT with sensors acquiring functional information like positron emission tomography (PET), single photon emission computed tomography (SPECT) or magnetic resonance spectroscopy (MRS) to study and analyze seizure disorders, Alzheimer's disease, depression and other diseases. Figure 1 shows a MEG-MRI co-registration, an example of Multi-Modal Registration.

Section 2 presents steps involved in Image registration, Section 3 contains classification criteria, Registration methods are presented in Section 4, Transform Model Estimation and Performance Analysis are discussed in Sections 5 and 6 respectively while Section 7 contains the conclusion.

---

[*]Corresponding author mail id: nagsayan112358@gmail.com

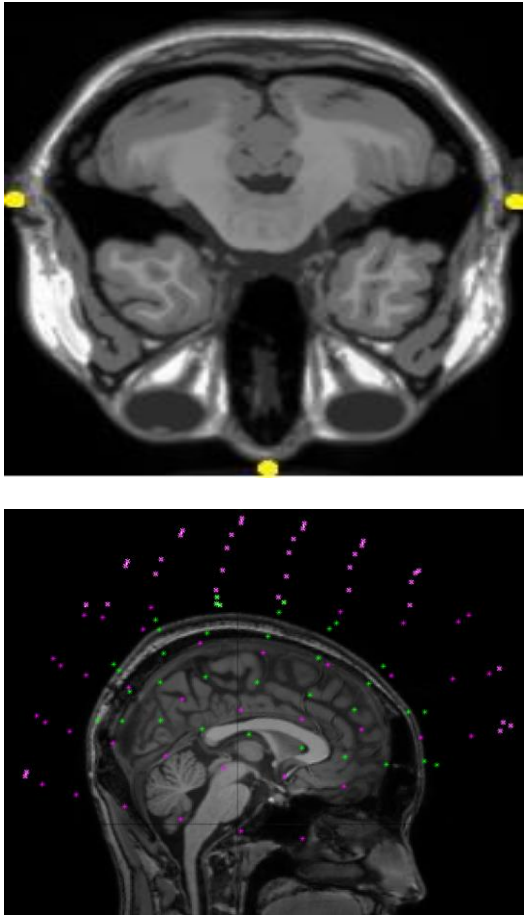

Fig. 1. Multimodal MRI-MEG Co-registration. Top-Yellow dots represent anatomical landmarks or fiducial points in the axial view of the brain image (anatomical information). Bottom- Pink dots represent the MEG sensors locations and the Green dots represent the scalp-EEG sensors locations. These MEG and EEG data contain the functional information and the bottom picture shows the co-registered brain image (sagittal view).

## II. STEPS INVOLVED IN IMAGE REGISTRATION

An Image Registration task involves the following steps as follows:

1. *Feature detection*: This is an important task of the Image Registration process. The detection process can be manual or automatic depending upon the complexity though automatic detection of features is preferred. Closed-boundary regions [9-16], edges, contours [17-26], line intersections, corners [27] along with their point representatives like center of gravity or line endings (collectively known as Control Points) can serve as features. These features consisting of distinctive objects must be easily detectable, that is, the features will be physically interpretable and identifiable. The feature set of the reference image must be sharing sufficient common features with the non-aligned image(s) irrespective of any undesired occlusions or unexpected changes for proper registration. The algorithm for detection should be robust enough to be able to detect the same features in all projections of the scene without being affected by any specific image deformation or degradation.

2. *Feature matching*: This step essentially establishes the correspondence between the features detected in the non-aligned sensed image and those detected in the reference image [28-36]. Different feature descriptors and similarity measures besides spatial relationships among the features are adopted to set up an accurate accordance. The feature descriptors must so formulated such that they remain unchanged in spite of any degradations and concurrently they must be able to properly discriminate among diverse features while remaining unaffected by noise.

3. *Transform model assessment*: For alignment of the sensed image with the reference image the parameters of the mapping functions are to be estimated [37-43]. These parameters are computed with the established feature correspondence obtained from the previous step. The selectivity of a mapping function depends on a priori knowledge regarding the acquisition process and expected image deformations. In absence of any a priori information the flexibility of the model must be ensured to tackle image deformations.

4. *Image transformation*: The sensed image is transformed for alignment employing the mapping functions.

The above mentioned image registration steps are generally followed. Figure 2 shows a pictorial representation of the steps involved in image registration. Though it is noteworthy to mention that it is difficult to fabricate a universal method applicable to all registration assignments the reason attributed to the diversity of images to be registered obtained from a miscellany of sources and the several types of degradations introduced in the images. Besides geometric deformation between the images, the radiometric deformations and noise corruptions should be taken into account for proper registration of images.

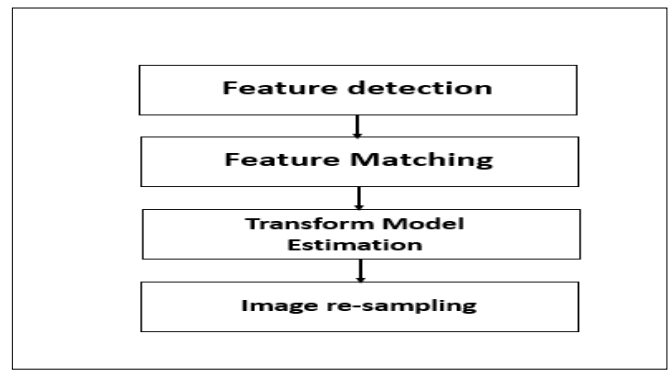

Fig. 2. Steps Involved in Image Registration

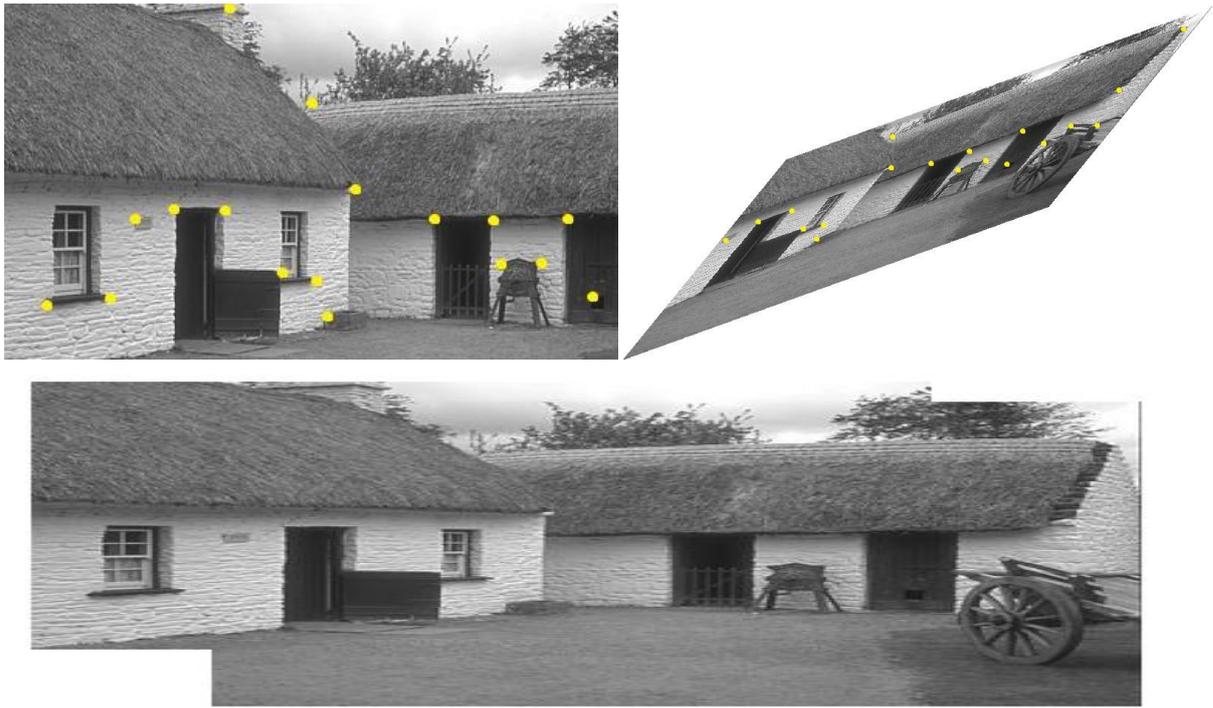

Fig. 3. Steps Involved in Image Registration. Top Left- Reference Image. Top Right- Non-Aligned Image. Bottom- Aligned Image. Yellow dots represent the extracted features and there are enough common features in both the images. A mapping function is established which gives the bottom image as the final output.

## III. CLASSIFICATION CRITERIA OF IMAGE REGISTRATION TECHNIQUES

Image registration techniques can be classified based on some criteria [44-45]. These are as follows:

1. *Dimensionality*: This specifies the dimensions of different possible registrations. It may be 2D-2D, 2D-3D or 3D-3D based on the requirement.
2. *Domain of transformation*: It may be global when the entire image is to be registered or it may be local when a portion of the image is taken into consideration for registration purpose.
3. *Type of transformation*: It may be rigid (translation, rotation, reflection), affine (translation, rotation, scaling, reflection, shearing), projective or non-linear.
4. *Registration Quality*: Depending on the data or the features extracted several measures can be adopted and applied.
5. *Parameters of Registration*: These are obtained employing search oriented methods. The optimum parameters found from a search method (e.g., a heuristic search method) determines the quality of transformation and hence the registration.
6. *Subject of Registration*: Same subject is considered for intra-subject registration. If the subjects are different then it is known as inter-subject registration.
7. *Object of Registration*: Different objects include head, abdomen, thorax, knee, etc.
8. *Nature of Registration basis*: It may be extrinsic (based on foreign objects which are easily detectable, e.g., markers glued to skin), intrinsic (based on image information) or non-image based (where imaging co-ordinates of the two devices are matched).
9. *Interaction*: It may be interactive, semi-automatic or entirely automatic.
10. *Modalities involved*: It may be mono-modal (which is also termed as intra-modal) using modalities like Computed Tomography (CT), Magnetic Resonance Imaging (MRI), Positron Emission Tomography (PET), Single Photon Emission Computed Tomography (SPECT), Ultra Sound (US), or Xray or Digital Subtraction Angiography (DSA) or multimodal (which is also known as inter-modal image) employing two or more modalities mentioned above.

## IV. METHODS OF IMAGE REGISTRATION

Various methods of Image Registration are as follows.

### 1. Extrinsic Methods

In this method artificial foreign objects which are easily detectable are attached to the patient body [46-53]. They serve as external features to be used for feature matching. The complexity is lessened and hence computational is fast and accuracy is also maintained. Examples are markers glued to patient's skin or stereo-tactic frame attached rigidly to the patient's outer skull for invasive neurosurgery related purposes.

## 2. Surface Methods

Surfaces or boundaries or contours are generally distinct in medical images unlike landmarks. For example, surface-based approach is employed for registering multimodality brain image. These surface matching algorithms are generally applied to rigid body registration. A collection of points, generally called a point set is extracted from the contours in an image. If two surfaces are considered for registration then there will be two such sets. The surface covering the larger volume of the patient, or that having a higher resolution if volume coverage is comparable, is generally considered for generation of the surface model. Iterative Closest Point Algorithm and Correspondence Matching Algorithm are successfully applied as registration algorithms for surface-based techniques [54-63]. Meta-heuristics and Evolutionary Optimization are also seen to solve these high dimensional optimization problems of surface registrations.

## 3. Moments and Principle Axes Methods

The orthogonal axes about which the moments of inertia are minimized are known as the principle axes. Two identical objects can be registered accurately by bringing their principal axes into concurrence without employing any rigid/affine transformations. If the objects are not identical but similar in appearance then they can be approximately registered by this technique [16, 64]. For moment based methods pre-segmentation is done in many cases to engender satisfactory outcomes.

## 4. Correlation Based Methods

This method is essentially useful for registration of mono-modal images and for comparison of several images of the similar object [65]. It has immense usage in the field of medical sciences for analyzing and treatment of disease. Extracted features from the images are also used to obtain the cross-correlation coefficients for image registration. [66-69]. Cross-correlation and Phase-correlation techniques based on Fourier domain are also used for image registration. Successful yet complex ventures have been significantly made using subspace-based frequency estimation approach for the Fourier based image registration problem employing multiple signal classification algorithm (MUSIC) to proliferate robustness eventually yielding accurate results [70]. Normalized mutual information between the images have been used for image registration purposes adopting an Entropy Correlation Coefficient (ECC) [71]. Fourier-based techniques accompanied by search algorithms have been exploited to evaluate the transformation between two input images [72].

## 5. Mutual Information Based Methods

In mutual information-based registration methods the joint probability of the intensities of comparable voxels in the images under consideration are estimated. Mutual information based measures are utilized to aid Voxel-based Registration. Mutual information can be fruitfully utilized for establishing the correspondence between the features of the reference and the sensed images as mentioned in the step of feature-matching. Correlation methods have proved inefficient for multi-modal registration. But, the mutual information based methods do not suffer from such a problem, rather they are found to perform effectively in multi-modal registration tasks. Gradient descent optimization methods have been employed to maximize mutual information [73]. Window and pyramid based approaches are used to achieve image registration using mutual information [74]. Other methods used include hierarchical search strategies along with simulated annealing [35] and the Powell's multi-dimensional direction set method [66]. Recently various optimization methods and multi-resolution strategies are adopted for mutual information maximization.

## 6. Wavelet Based Methods

Wavelet Transform was introduced to get an idea of the time instant at which a particular frequency exists. The width of the window is altered as the transform is computed for each spectral component- the most important characteristic of the multi-resolution wavelet transform. It offers both time and frequency selectivity, that is, it is able to localize properties in both temporal and frequency domains. The wavelet-based image registration can be effectively. After choosing several wavelet coefficients by selection rules like the maximum absolute wavelet coefficient in the multi-spectral image and the high-resolution image for individual band the partial wavelet coefficients of the high-resolution image are replaced with those of the multi-spectral low-resolution image. The pyramidal approaches also use wavelet decomposition owing to its intrinsic multiresolution properties. Different types of wavelets like the Haar, Symlet, Daubechies [75] and Coiflets are applied for finding the correspondence with different sets of wavelet coefficients. Wavelet-based feature extraction techniques along with normalized cross-correlation matching and relaxation-based image matching techniques are used thereby incorporating sufficient control points to reduce the local degradations, for image registration [76].

## 7. Soft Computing Based Methods

These methods are comparatively recent and advanced and are successfully applied to image registration tasks. They include Artificial Neural Networks, Fuzzy Sets and several Optimization Heuristics.

a) *Artificial Neural Networks:*

An artificial neural network (ANN) is a computational model which is formulated based on biological neural networks. It is also known as Multi-Layer Perceptron (MLP) since it contains a number of hidden layers. These layers consist of an interconnected group of artificial neurons an information is passed on from one layer to the next layer. Artificial Neural Networks or simply Neural Networks learns adaptively in the learning phase when information flows through the network and updates the neuron-links accordingly by assigning various weights to them. Neural Networks can be viewed upon as non-linear statistical data modeling tools employed to model complex relationships between inputs and outputs or to recognize patterns in data, also called Pattern Recognition. There are two types of schemes: (1) feed-forward networks, where the links are devoid of any loop (e.g., multilayer perceptron (MLP) and radial basis function neural networks (RBF) and (2) recurrent networks which include loops (e.g., self-organizing maps (SOM) and Hopfield Neural Networks). A priori information about the output is an essential requirement for training feed forward networks, on the other hand, recurrent neural networks generally do not require any such previous knowledge regarding the expected output. The rigorous training process in an ANN modifies and adaptively updates the network architecture abreast the connection weights or link weights to be able to learn complex non-linear input-output relationships thereby parlaying the robustness and efficacy of performance. Multi-layer Perceptron, Radial basis functions, self-organizing maps and Hopfield networks have been utilized for different computational and optimization aspects and for designing registration matrices in Image Registration problems [77]. Neural Networks have also been used for solving mono-modal and multi-modal medical image registration problems [78].

*b)  Fuzzy Sets:*

A fuzzy set is a collection of elements having a continuous sequence of membership grades or degrees. Fuzzy sets was introduced by L. A. Zadeh in 1965. Fuzzy sets follow the properties of inclusion, union, complement, intersection, etc. In classical set theory, the membership values of elements in a set are decided in binary terms depending upon whether an element belongs or does not belong to the set. In contrast, fuzzy set theory allows the grading of the membership of elements in a fuzzy set as decided with the assistance of a membership function which assigns values residing in the interval [0, 1]. Fuzzy Sets manifest the perception of partial membership of an element within the set- this permits Fuzzy sets to tackle uncertainty and inaccuracies. Fuzzy Sets have been explicitly applied to Image registration techniques [79-80]. It has also been utilized to choose and pre-process the extracted features to be registered. Fuzzy logic is used to enhance the precision in the transformation parameters as estimated approximately previously eventually leading to accurate registration estimates [81].

*c)  Optimization Heuristics:*

Optimization problems applied in several domains of Engineering Design and Optimization have some mathematical models and objective functions. They may be unconstrained (without constraints) or constrained (with constraints) having both continuous as well as discrete variables. The task of finding the optimal solutions is difficult with numerous curtailments being active at the points of global optima. Traditional methods including Gradient Descent, Dynamic Programming and Newton Methods are computationally less efficient, whereas provide feasible solutions in a stipulated time. The list of meta-heuristics include Genetic Algorithm (GA) [82], Particle Swarm Optimization (PSO) [83], Gravitational Search Algorithm (GSA) [84], Ant Colony Optimization (ACO) [85-86], Stimulated Annealing (SA) [87-88], and Plant Propagation Algorithm (PPA) [89-90] and so on.

GA is a relatively old, approximate search technique used in computing. These global search heuristics form an important class of evolutionary algorithms that mimics evolutionary biological processes such as mutation, selection, and crossover and abandonment. Likewise, Particle Swarm Optimization and Differential Evolution along with their existing variants are relatively advanced heuristics than can efficiently solve Optimization problems. These optimization heuristics are applied to image registration problems for finding the optimal parameters necessary for designing a transformation model [91].

V. TRANSFORM MODEL ESTIMATION

A transformation is expounded as the process of mapping a set of points to various other locations. The objective is to design a proper transformation model which transforms the sensed image with respect to the original image with maximum accuracy. The transformations that may be performed are translation, rotation, scaling, shearing and reflection. These are collectively known as affine transformation. Also there are projective and non-linear transformations as well.

*1. Translation*

Let a point $x$ is to be translated by $t$ units, then the matrix representation of this transformation is given as:

$$\begin{bmatrix} y_1 \\ y_2 \end{bmatrix} = \begin{bmatrix} x_1 \\ x_2 \end{bmatrix} + \begin{bmatrix} t_1 \\ t_2 \end{bmatrix} \quad (1)$$

where, $y_1$, $y_2$ = new point, $x_1$, $x_2$ = old point, $t_1$, $t_2$ = translation value.

*2. Rotation*

A point with co-ordinate $P_1(x_1, x_2)$ on a 2-D plane is rotated by an angle $\theta$ with respect to origin then the relationship between the final point $P_2(y_1, y_2)$ and the initial point is given as:

$$\begin{bmatrix} y_1 \\ y_2 \end{bmatrix} = \begin{bmatrix} cos\theta & sin\theta \\ -sin\theta & cos\theta \end{bmatrix} + \begin{bmatrix} x_1 \\ x_2 \end{bmatrix} \quad (2)$$

where, $y_1$, $y_2$ = new point, $x_1$, $x_2$ = old point, $\theta$ = rotational parameter.

*3. Scaling*

Scaling is required to resize an image, or to work with images whose voxel sizes differ between images. It is represented as:

$$\begin{bmatrix} y_1 \\ y_2 \end{bmatrix} = \begin{bmatrix} s_1 & 0 \\ 0 & s_2 \end{bmatrix} + \begin{bmatrix} x_1 \\ x_2 \end{bmatrix} \quad (3)$$

where, $y_1$, $y_2$ = new point, $x_1$, $x_2$ = old point, $s_1$, $s_2$ = scaling parameters.

*4. Shearing*

In shearing the parallel lines are only preserved. It may be represented as:

$$\begin{bmatrix} y_1 \\ y_2 \end{bmatrix} = \begin{bmatrix} a_{11} & a_{12} \\ a_{21} & a_{22} \end{bmatrix} \begin{bmatrix} x_1 \\ x_2 \end{bmatrix} + \begin{bmatrix} a_{13} \\ a_{23} \end{bmatrix} \quad (4)$$

where, $y_1$, $y_2$ = new point, $x_1$, $x_2$ = old point, $a_{11}$, $a_{12}$, $a_{13}$, $a_{21}$, $a_{22}$, $a_{23}$, = shearing parameters. Fig. 4. shows an example of shearing transformation.

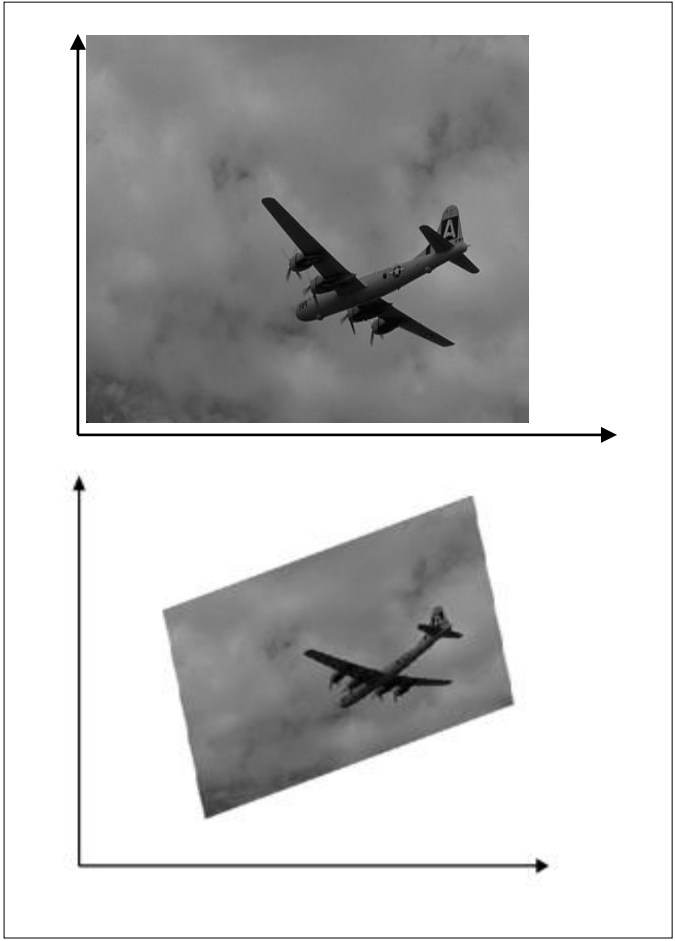

Fig. 4. Example of Affine Transformations (Translation, Rotation, Scaling and Shearing)

## VI. PERFORMANCE ANALYSIS

It is required to estimate how accurate the registration actually is. Also to qualitatively analyze the performance of the algorithms some metrics are used. They also serve as the basis for improvement in the registration in each iteration. The selection of similarity measures depends on modality of images to be registered. Correlation based metrics like Correlation Coefficient is applicable to mono-modal registration and Mutual Information is utilized for multi-modal image registration purposes.

*a) Correlation Coefficient (CC):*

CC is essentially a similarity measure which gives an idea of how well the reference and transformed images are identical [32-34]. If two images are perfectly identical, CC gives a value equal to 1, whereas, if the two images are completely uncorrelated CC value is equal to 0 and CC value equal to -1 indicates that the images are completely anti-correlated, which means one image is the negative of the other. It gives satisfactory results with mono-modal registration. It is represented as:

$$CC = \frac{\sum_i (x_i - x_m)(y_i - y_m)}{\sqrt{\sum_i (x_i - x_m)^2} \sqrt{\sum_i (y_i - y_m)^2}} \quad (5)$$

where $x_i$, $y_i$ = intensity of $i^{th}$ pixel in the reference and sensed image respectively, and $x_m$, $y_m$ = mean intensity of reference and sensed image respectively.

*b) Mutual Information (MI):*

MI is yet another measure determining the degree of similarity measured between the image intensities of corresponding voxels in both images [35-36]. MI is maximized when both the images are accurately aligned. The values of MI are non-negative and symmetric. The range of MI values starts from zero and can vary up to a high value. High MI value depicts large reduction in uncertainty whereas zero MI value is clear indication that the two variables are independent. It is represented as:

$$MI(x, y) = \sum_{y \in Y} \sum_{x \in X} p(x, y) \log \left( \frac{p(x, y)}{p_1(x) p_2(y)} \right) \quad (6)$$

where $p(x, y)$ = joint distribution function and $p_1(x)$, $p_2(y)$ = marginal distribution functions.

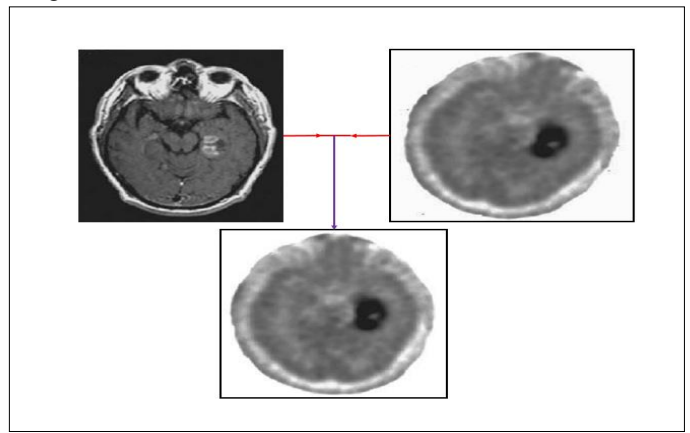

Fig. 5. Example of Multi-modal Image registration using Mutual Information as similarity measure. Top Left- Reference MRI brain image (axial view), Top

Right- Non- Aligned PET brain image (axial view), Bottom- Transformed PET brain image (axial view).

VII. CONCLUSION

This paper tries to present a survey of the registration methods along with the detailed classifications among various approaches. Image registration is an essential step for integrating or fusing and analyzing information from various sensors (sources). It has immense applications in fields of medical sciences, computer vision and remote sensing. Image registrations with complex nonlinear distortions, multi-modal registration and registrations of occluded images despite being affected by illumination factors among others thus contributing to the robustness of the approaches belong to the most challenging tasks at the present scenario. Generation of features or control points and the mapping or transformation functions are essential steps and a lot of research work needs to be done to enhance the accuracy. In multimodal registration, MI technique has gained popularity in particular whereas for mono-modal images correlation based similarity metrics are preferred. Robustness and Reliability can be proliferated by hybrid approaches combining MI based techniques with feature-based measures. Several soft computing methods including the optimization heuristics are applied to find the optimum parameters mostly in case of affine transformations based registration. No gold standard algorithms or approaches can be developed for image registration purposes because of the dependency on the images under consideration. Thus, despite a lot of work has been done, automatic image registration is still considered as an open problem. The future works will be introducing new feature-based methods, where apt modality-insensitive features can provide robust as well as accurate outcomes for the registration.

ACKNOWLEDGMENT

I would like to extend my sincere gratitude to Professor Sugata Munshi, Professor Amitava Chatterjee and Professor Mita Dutta for their support and guidance.